\documentclass[runningheads]{llncs}
\usepackage[hidelinks]{hyperref} 
\usepackage[T1]{fontenc}
\usepackage{graphicx} 
\usepackage{amsmath}
\usepackage{arydshln}
\usepackage{tabularray}
\usepackage{adjustbox}
\usepackage{multirow}
\usepackage{booktabs}
\usepackage{float}
\usepackage{algorithm}%
\usepackage{algorithmicx}%
\usepackage{cite}
\usepackage[table]{xcolor}
\usepackage{colortbl}
\definecolor{lightgray}{gray}{0.9}
\usepackage{orcidlink}

\usepackage{enumitem}
\setlist{nosep}
\begin{document}
	\title{Minimalist Preprocessing Approach for Image Synthesis Detection} 
%
%
\author{Hoai-Danh Vo\inst{1,2}\orcidlink{0009-0006-6774-2123}, Trung-Nghia Le\thanks{Corresponding author.}\inst{1,2}\orcidlink{0000-0002-7363-2610}}
%
\authorrunning{H.-D. Vo and T.-N. Le}
%
\institute{University of Science, VNU-HCM, Ho Chi Minh City, Vietnam \and
	Vietnam National University, Ho Chi Minh City, Vietnam\\
	\email{22c15025@student.hcmus.edu.vn, ltnghia@fit.hcmus.edu.vn}}
%
\maketitle              
\begin{abstract}
	Generative models have significantly advanced image generation, resulting in synthesized images that are increasingly indistinguishable from authentic ones. However, the creation of fake images with malicious intent is a growing concern. Low-configured smart devices have become highly popular, making it easier for deceptive images to reach users. Consequently, the demand for effective detection methods is increasingly urgent. 
	In this paper, we introduce a simple yet efficient method that captures pixel fluctuations between neighboring pixels by calculating the gradient, which highlights variations in grayscale intensity. This approach functions as a high-pass filter, emphasizing key features for accurate image distinction while minimizing color influence.
	Our experiments on multiple datasets demonstrate that our method achieves accuracy levels comparable to state-of-the-art techniques while requiring minimal computational resources. Therefore, it is suitable for deployment on low-end devices such as smartphones. The code is available at \href{https://github.com/vohoaidanh/adof}{https://github.com/vohoaidanh/adof.}
	
\end{abstract}

\section{Introduction}
%

In recent years, significant advancements in image generation have been achieved, particularly with Generative Adversarial Networks (GANs)~\cite{Goodfellow2014GenerativeAN} and Diffusion models~\cite{Ho2020DenoisingDP,Gu2021VectorQD}. These approaches produce high-quality images that closely resemble real-world visuals~\cite{spottingai} and have garnered attention in academic and societal circles. Generative models have found applications in various fields, including virtual try-ons and personalized fashion recommendations in the fashion industry~\cite{8769486}, as well as in image editing~\cite{CasteleiroPitrez2024GenerativeAI,Shin2024CanPM} and interior design~\cite{Chen2020ApplicationOA}.

\begin{figure}[t!]
\centering
\includegraphics[width=0.7\textwidth]{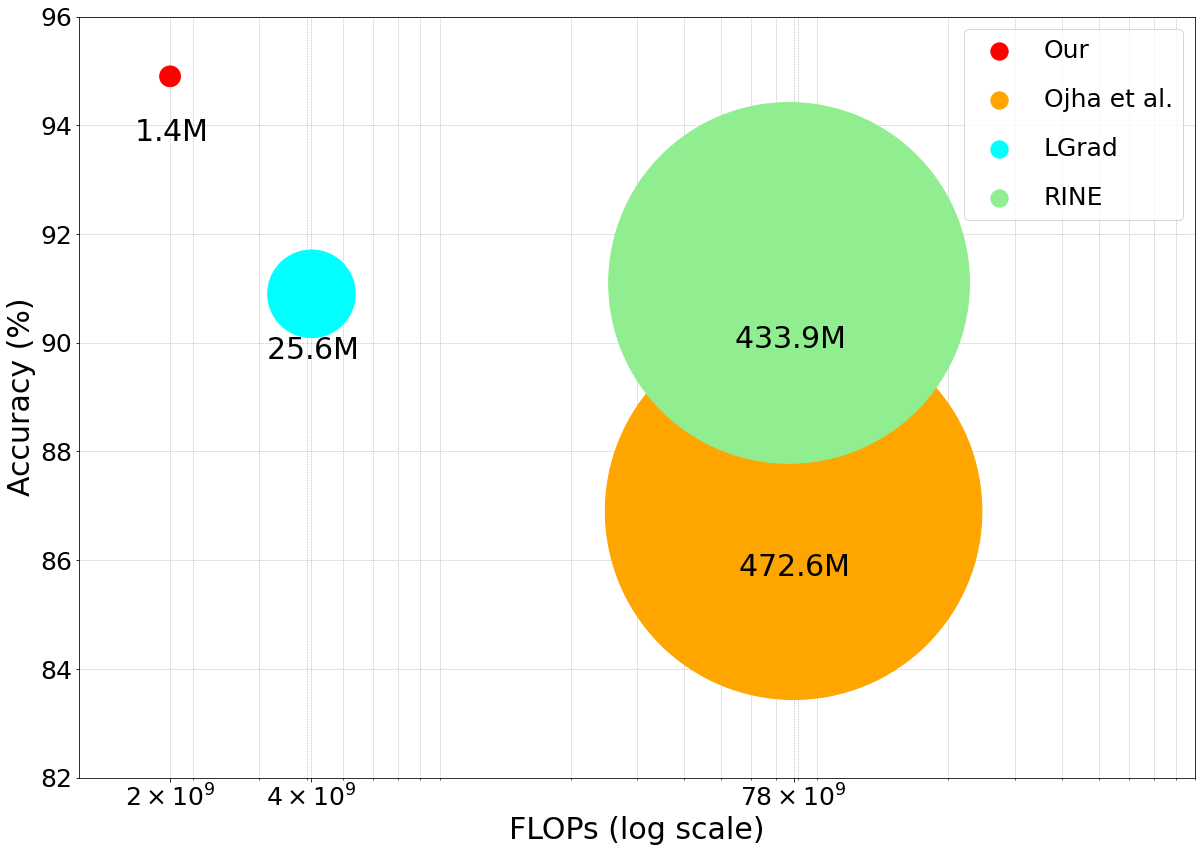}
\caption{Comparison of different synthetic image detection methods on the Ojha dataset~\cite{Ojha2023TowardsUF}.
	Our proposed method is simple yet efficient, significantly reduces FLOPs, total parameters, while achieving comparable accuracy state-of-the-art methods.}
\label{fig:teaser}
\end{figure}
Despite the valuable applications of image generation technology, significant drawbacks exist. According to a survey conducted by Bauer and Bindschaedlerr~\cite{DBLP-abs-2107-10139}, generative models can create fake information, particularly deepfakes, which depict fabricated scenarios involving famous individuals. In response to these dangers, several US states~\cite{CaliforniaDeepfakes,VirginiaDeepfake} have outlawed the malicious use of deepfake technology, especially for harmful content like revenge and celebrity pornography.
\textcolor{black}{To address the threats posed by synthetic images on digital communication platforms and social media, it is essential to develop effective countermeasures for verifying image authenticity directly on mobile devices. Given the ubiquity and portability of these devices, real-time detection of generated images is crucial for preventing misinformation and preserving the integrity of visual content. However, the constrained computational capacity of mobile devices presents a significant challenge. This paper introduces a simple yet efficient solution for synthesized image detection, specifically the Adjacency Difference Orientation Filter (ADOF) for data preprocessing;}
\textcolor{black}{this filter allows us to compute the gradient in both the $x$ and $y$ directions. The direction of the gradient reflects the behavior of grayscale variation among neighboring pixels, assisting in distinguishing between real and generated images. Focusing on extracting useful low-level features, our approach ensures generalization while utilizing a lightweight CNN architecture for detecting generated images, without demanding extensive computational resources. This strategy effectively reduces irrelevant information, enabling the model to concentrate on fine-grained variations, ultimately leading to improved performance and generalization.}
\textcolor{black}{In contrast to existing methods \cite{koutlis2024leveraging,Ojha2023TowardsUF,Tan2023LearningOG} that require large deep learning architectures, such as CLIP~\cite{abs-2103-00020}, ViT~\cite{abs-2010-11929}, Resnet50~\cite{He2015DeepRL}, and significant computational resources, our approach demands fewer resources while still ensuring generalization and achieving comparable accuracy. Fig. \ref{fig:teaser} presents a comparative overview of results, highlighting the advantages of this strategy. }

Experiments on well-known datasets~\cite{Ojha2023TowardsUF,Wang2023DIREFD,Wang2019CNNGeneratedIA} demonstrate the effectiveness of our method, achieving impressive accuracy of 94.9\% on the Ojha dataset~\cite{Ojha2023TowardsUF} and 98.3 \% on the DiffusionForensis~\cite{Wang2023DIREFD}. Additionally, there is a  reduction in computational load by 97.8\% compared to RINE~\cite{koutlis2024leveraging} and 57.8\% compared to LGrad~\cite{Tan2023LearningOG}. These results underscore the advantages of our approach, as illustrated in Fig. \ref{fig:teaser}. The code for reproducing our results is publicly available at \href{https://github.com/vohoaidanh/adof}{https://github.com/vohoaidanh/adof.} Our contributions are as follow:
\begin{itemize}
\item \textcolor{black}{We introduce a simple yet efficient approach for detecting synthetic images, and our approach is more generalized than existing methods.}

\item \textcolor{black}{We present a filter-based method that computes pixel intensity gradients to capture pixel fluctuations and reduce color influence, leading to improved model performance with faster inference speed and lower complexity.} 

\item \textcolor{black}{Our proposed method reduces the number of parameters and FLOPs while maintaining accuracy compared to state-of-the-art methods.}
\end{itemize}

\section{Related Work}

%
Various methods have been developed to address the challenge of distinguishing synthetic images from real ones, utilizing both traditional machine learning techniques and modern deep learning approaches.
Durall \textit{et al.}~\cite{Durall2019UnmaskingDW} applied a Fourier Transform~\cite{Arunachalam2013TheFF} to grayscale images and used azimuthal averaging to convert the 2D frequency data into a 1D feature vector, retaining essential information for classification. They then employed either Support Vector Machines or K-means clustering to detect GAN-generated images.
Alternatively, methods like RINE~\cite{koutlis2024leveraging} and Ojha et al.~\cite{Ojha2023TowardsUF}, along with similar approaches, leverage pre-trained deep learning networks such as CLIP to enhance performance. This integration contributing to consistently high success rates in detecting synthesized images through the integration of these networks into their frameworks.
Notably, the FatFormer~\cite{Liu_2024_CVPR} method focuses on the contrastive objectives between adapted image features and text prompt embeddings, providing valuable information that enables the deep learning models to learn more robust and discriminative representations, ultimately improving their ability to accurately classify real and generated images.

\textbf{Frequency domain-based methods} involve transforming images from the spatial domain to the frequency domain using transformations such as Fast Fourier Transform or Discrete Cosine Transform (DCT). By focusing on frequency characteristics, these methods effectively capture artifacts that might not be evident in the spatial domain. This allows classifiers to distinguish between real and fake images by analyzing the unique patterns that emerge in the spectral domain. 
Frank \textit{et al.}~\cite{Frank2020LeveragingFA} utilized the DCT to analyze images in the frequency domain, revealing unique spectral differences between real images and those produced by GAN models.
Qian \textit{et al.} introduced $F^3$-\textit{Net}~\cite{Qian2020ThinkingIF} \textcolor{black}{to decompose the spectrum into various bands, enabling the analysis of these components to identify unusual distributions. This method effectively detects subtle artifacts, enhancing the ability to recognize synthetic image manipulations.}
%
%

%
%
Tan \textit{et al.} proposed FreqNet~\cite{Tan2024FrequencyAwareDD}, which emphasizes high-frequency details and directs the detector to concentrate on these features across spatial and channel dimensions, rather than utilizing the full spectrum of frequency bands as is common in many other approaches.
\textcolor{black}{BiHPF~\cite{Jeong2021BiHPFBH}, a method by Jeong \textit{et al.}, amplifies frequency-level artifacts commonly found in images generated by generative models to tackle the challenge of identifying images from previously unobserved models.}
%
Jeong \textit{et al.}~\cite{Jeong2022FrePGANRD} generated perturbation maps added to training images to prevent overfitting to frequency-specific features, reducing high-frequency noise and enhancing classifier generalization.
%
%

\textbf{Spatial-based methods} analyze images directly on pixel values, as seen in models like CNNDetection~\cite{Wang2019CNNGeneratedIA} and Gram-Net~\cite{9157447}. A key issue, however, is that raw images often contain excessive, irrelevant information, such as semantic content, which Nang \textit{et al.}~\cite{zhong2024patchcraftexploringtexturepatch} identified as detrimental to image classification effectiveness. This extraneous information, unnecessary for distinguishing real from fake images, can disrupt the model’s learning process and reduce its effectiveness.
%
%
%
Wang \textit{et al.}~\cite{Wang2019CNNGeneratedIA} developed a comprehensive detector to distinguish real images from CNN-generated ones~\cite{Krizhevsky2012ImageNetCW}. Using a dataset of images from 11 CNN-based generators, they showed that, with effective pre-/post-processing and data augmentation, a classifier trained on ProGAN~\cite{Karras2017ProgressiveGO} generalizes well to other models, including StyleGAN2~\cite{Karras2019AnalyzingAI}.
%
%
%
%
\textcolor{black}{By dividing images into small patches categorized as either rich texture or flat, PatchCraft~\cite{zhong2024patchcraftexploringtexturepatch} exploits the inter-pixel correlation contrast between these regions. This approach breaks the semantic coherence present in traditional methods, addressing a key limitation and enhancing the model's ability to generalize more effectively.
}
Tan \textit{et al.} introduced the concept of Neighboring Pixel Relationships (NPR)~\cite{Tan2023RethinkingTU} to capture and characterize generalized structural artifacts that arise from up-sampling operations, which are commonly used in image generation models to enhance image quality. This method shows a significant improvement over other techniques within the same approach.

%
%
%
Frequency-based approaches achieve faster convergence with smaller models but may lose essential spatial information needed to distinguish real from generated images. In contrast, spatial-based methods often require larger models and may struggle with new domain data.
Our method leverages the strengths of both approaches by focusing on pixel perturbations between neighboring pixels, effectively discarding much of the pixel color information. Additionally, our filter functions as a high-pass filter, removing low-frequency components to emphasize the relevant features.
\section{Proposed Method}
\begin{figure}[ht!]
\centering
\includegraphics[width=0.8\textwidth]{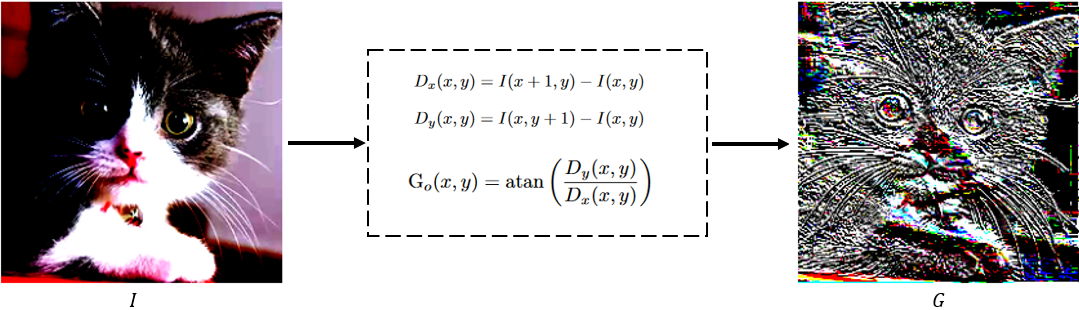}
\caption{The left image represents the original image, the middle shows the gradient calculation applied, and the right image illustrates the resulting gradient map.}
\label{figure_cat_filter}
\end{figure}

\subsection{Overview}

Generative models, such as GAN~\cite{Goodfellow2014GenerativeAN} and Diffusion~\cite{Ho2020DenoisingDP}, currently use CNN layers for image synthesis, meaning that neighboring pixel regions are correlated to a certain extent.
We hypothesize that synthetic images exhibits a stronger correlation between adjacent pixels compared to real images.
Furthermore, due to the design of neural networks, noise in synthetic images tends to be averaged out, whereas in real images, noise typically remains more prominent.
To investigate and evaluate the impact of noise on adjacent pixels, we designed a simple yet effective filter to capture these variations.
This filter aids the CNN in learning the differences in noise distribution between real and synthetic images.
The overall architecture of ADOF is illustrated in Fig. \ref{figure_cat_filter}. This approach captures noise information by calculating the differences between adjacent pixels and incorporating these differences into the gradient to account for variations in both the \(x\) and \(y\) directions. 

\subsection{Adjacency Difference Orientation Filter (ADOF)}

\subsubsection{Finite Difference.} 
The finite difference technique~\cite{Mickens2015DifferenceE,finite_difference} is a mathematical approach for estimating intensity variations across neighboring points in a grid or matrix. In image processing, it calculates gradients by measuring differences in pixel values, thereby detecting changes in intensity in both horizontal and vertical directions. This technique aids in edge detection and texture analysis by highlighting contrasts between adjacent pixels.

The general formula for calculating the gradient in a given direction \( u \) is \( \text{Gradient}_{u} = I(x + \Delta x, y + \Delta y) - I(x, y) \), where \( \Delta x \) and \( \Delta y \) define the direction of the difference.







\subsubsection{Filter Construction. } 
We have applied the \textit{finite difference} to compute the gradients of intensity in our images. This helps in identifying intensity variations at each pixel and supports the detection of important geometric features in our approach. The formula that computes the difference between adjacent pixels along the \(x\) and \(y\)-direction is given by:
\begin{equation}
D_x(x, y) = I(x+1, y) - I(x, y),
\end{equation}
\begin{equation}
D_y(x, y) = I(x, y+1) - I(x, y),
\end{equation}
where \(I\) represents the image in which the difference is being calculated, \(D_x(x, y)\) represents the difference between the pixel value at \((x+1, y)\) and the pixel value at \((x, y)\). This filter captures variations in pixel intensity along the horizontal direction. Similarly, \(D_y(x, y)\) captures variations in pixel intensity along the vertical direction.
To determine the gradient magnitude and orientation, these values are computed from \(D_x\) and \(D_y\).
\begin{equation}
\text{G}_m(x, y) = \sqrt{D_x(x, y)^2 + D_y(x, y)^2},
\end{equation}
\begin{equation}
\text{G}_o(x, y) = \text{arctan} \left( \frac{D_y(x, y)}{D_x(x, y)} \right),
\end{equation}
where \(D_x(x, y)\) and \(D_y(x, y)\) are as previously defined. The gradient orientation \(\text{G}_o\), which represents the overall angle of gray-level changes at a pixel and indicates the direction of these combined intensity variations, is referred to as the \textbf{\textit{Adjacency Difference Orientation Filter (ADOF)}} in this paper. Meanwhile, the gradient magnitude \(\text{G}_m\) quantifies the strength of intensity changes at that pixel. The result of this computational process is illustrated in Figure~\ref{figure_cat_filter}.

\subsection{Lightweight Model Architecture}
\begin{figure}[t!]
	\centering
	\includegraphics[width=0.8\textwidth]
	{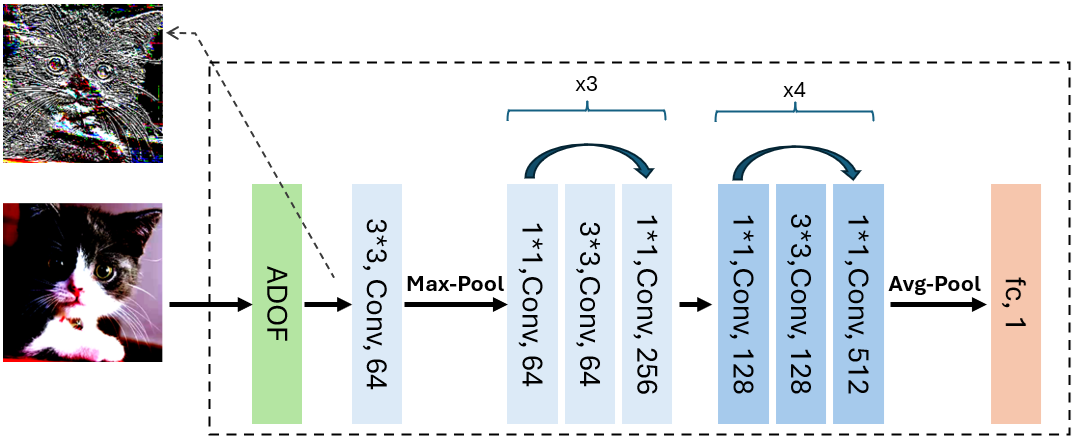}
	\caption{Architecture of our lightweight model.}
	\label{figure_model_architecture}
\end{figure}

To evaluate the effectiveness of our filter ADOF on images, we use basic CNN architectures. Specifically, this work employs a modified ResNet50~\cite{He2015DeepRL} model with \texttt{layer3} and \texttt{layer4} removed. To capture information from 8-connected neighboring pixels more effectively, the kernel size of the \texttt{conv1} layer was adjusted from 7 to 3. The architecture is depicted in Fig.~\ref{figure_model_architecture}.
\section{Experiments}

\subsection{Implementation Details}


In practice, we are more concerned with the flat regions of an image rather than the edge areas where there is a significant variation in gray levels between the x and y directions. This is because, in regions with large changes in gray levels in one direction compared to the other, the gradient angles are close to \(\pm \frac{\pi}{2}\). Although these angles both indicate edge regions in the image, the gradient angles at edges typically take values of \(\pm \frac{\pi}{2}\), which are numerically distant from each other despite conveying similar edge information. To exclude these areas, we set the gradient values approaching \(\pm \frac{\pi}{2}\) to 0 
, the experimental process has demonstrated that this approach leads to higher accuracy for the model.

All experiments are conducted on a computing system using a NVIDIA RTX A4000 GPU with 16 GB of memory and an AMD Ryzen 5 5600X 6-Core CPU. We trained our model using parameters that are closely aligned with those used in common methods~\cite{Wang2019CNNGeneratedIA,Tan2023RethinkingTU,Tan2023LearningOG} to ensure a fair comparison and demonstrate the effectiveness of our method independent of specific hyperparameters. Furthermore, we utilized the source code provided by NPR~\cite{Tan2023RethinkingTU} to streamline the training process and maintain consistency. The model was trained using the Adam optimizer with a learning rate of $2 \times 10^{-4}$ and a batch size of 32. To accelerate the training process, we adjusted the learning rate every 5 epochs instead of every 10 epochs \textcolor{black}{and utilized 4 out of the 20 classes (\textit{car, cat, chair, horse}) for training, similar to the protocol used in existing works~\cite{Wang2019CNNGeneratedIA,Jeong2022FrePGANRD,Tan2023RethinkingTU,Jeong2021BiHPFBH}.  
}


\subsection{Dataset}
\subsubsection*{Training set.} To facilitate comparison between methods, we used the same ForenSynths dataset with existing methods~\cite{Wang2019CNNGeneratedIA,Ojha2023TowardsUF,Tan2023RethinkingTU,Tan2023LearningOG,9897820}. This dataset consists of 20 object classes selected from the LSUN dataset. Each class contains 18,000 real-world images, with corresponding generative images generated using the ProGAN~\cite{Karras2017ProgressiveGO} model. To verify the generalization of methods, all compared method was trained on a subset of the ForenSynths~\cite{Wang2019CNNGeneratedIA} dataset consisting of 4 classes: car, cat, chair, horse. 
\subsubsection*{Evaluation set.}
%
To investigate the generalization of methods, our evaluation was conducted using the Self-Synthesis 9~GANs~\cite{Tan2023RethinkingTU}, which contains 36,000 images sourced from LSUN, ImageNet, CelebA, CelebA-HQ, COCO, and FaceForensics++, generated using models like AttGAN, BEGAN, and CramerGAN. The second dataset, DiffusionForensics~\cite{Wang2023DIREFD}, comprises 40,000 images from LSUN and ImageNet, utilizing models such as ADM, DDPM, and IDDPM. Lastly, the Ojha Test Set~\cite{Ojha2023TowardsUF} includes 16,000 images from LAION and ImageNet, generated with ADM, Glide, DALL-E-Mini, and LDM.

\subsection{Comparison with State-of-the-Art Methods}
We conduct a performance comparison of our method with 10 State-of-the-Art methods, including 
CNNDetection~\cite{Wang2019CNNGeneratedIA}, 
Frank~\cite{Frank2020LeveragingFA}, 
Durall~\cite{Durall2020WatchYU}, 
Patchfor~\cite{Chai2020WhatMF}, 
F3Net~\cite{Qian2020ThinkingIF} , 
SelfBland~\cite{Shiohara2022DetectingDW}, 
GANDetection~\cite{Mandelli2022DetectingGI}, 
LGrad~\cite{Tan2023LearningOG} , 
Ojha~\cite{Ojha2023TowardsUF}, 
NPR~\cite{Tan2023RethinkingTU}. The experimental results in Tables~\ref{tab:table2}, \ref{tab:table3}, and \ref{tab:table4} demonstrate that our method exceeds existing approaches. 
\textcolor{black}{On the 9-GAN dataset, ADOF delivers the highest accuracy at 94.2\%, surpassing Ojha~\cite{Ojha2023TowardsUF} with a mere 77.6\% \ref{tab:table2}, and NPR~\cite{Tan2023RethinkingTU} at 93.2\% (see Table~\ref{tab:table2}).}
\textcolor{black}{Notably, our approach achieves a remarkable 98.3\% accuracy on the DiffusionForensics~\cite{Wang2023DIREFD} dataset, outperforming the NPR method~\cite{Tan2023RethinkingTU}, which only reaches 95.3\% (see Table~\ref{tab:table3}). It also surpasses DIRE~\cite{Wang2023DIREFD}, which reports 97.9\% accuracy on its own dataset, despite our model being trained on Forensynths~\cite{Krizhevsky2012ImageNetCW}, in contrast to DIRE’s training on DiffusionForensics. Additionally, this approach exceeds both RINE~\cite{koutlis2024leveraging} and Ojha~\cite{Ojha2023TowardsUF} (see Table~\ref{tab:table4}), with the latter achieving 91.1\% on its dataset~\cite{Ojha2023TowardsUF}. It is worth mentioning that both methods utilize a large CLIP model for their evaluations.}

\begin{table}[t!]
\centering
\caption{Evaluation results on the Self-Synthesis 9~GANs~\cite{Tan2023RethinkingTU}.}
\label{tab:table2}
\begin{adjustbox}{max width=\textwidth}
\begin{tabular}{l cc cc cc cc cc cc cc cc cc|cc}
\hline
\multirow{2}{*}{{\textbf{Method}}} & \multicolumn{2}{c}{\textbf{AttGAN}} & \multicolumn{2}{c}{\textbf{BEGAN}} & \multicolumn{2}{c}{\textbf{CramerGAN}} & \multicolumn{2}{c}{\textbf{InfoMaxGAN}} & \multicolumn{2}{c}{\textbf{MMDGAN}} & \multicolumn{2}{c}{\textbf{RelGAN}} & \multicolumn{2}{c}{\textbf{S3GAN}} & \multicolumn{2}{c}{\textbf{SNGAN}} & \multicolumn{2}{c|}{\textbf{STGAN}} & \multicolumn{2}{c}{\textbf{Mean}} \\
\cline{2-21}
 & {\textbf{Acc.}} & {\textbf{A.P.}} & {\textbf{Acc.}} & {\textbf{A.P.}} & {\textbf{Acc.}} & {\textbf{A.P.}} & {\textbf{Acc.}} & {\textbf{A.P.}} & {\textbf{Acc.}} & {\textbf{A.P.}} & {\textbf{Acc.}} & {\textbf{A.P.}} & {\textbf{Acc.}} & {\textbf{A.P.}} & {\textbf{Acc.}} & {\textbf{A.P.}} & {\textbf{Acc.}} & {A.P.} & {\textbf{Acc.}} & {\textbf{A.P.}} \\
\hline
CNNDetection~\cite{Wang2019CNNGeneratedIA}  & 51.1 & 83.7 & 50.2 & 44.9 & 81.5 & 97.5 & 71.1 & 94.7  & 72.9 & 94.4 & 53.3 & 82.1 & 55.2 & 66.1 & 62.7 & 90.4 & 63.0 & 92.7 & 62.3 & 82.9 \\
Frank~\cite{Frank2020LeveragingFA} & 65.0 & 74.4 & 39.4 & 39.9 & 31.0 & 36.0 & 41.1 & 41.0 & 38.4 & 40.5 & 69.2 & 96.2 & 69.7 & 81.9 & 48.4 & 47.9 & 25.4 & 34.0 & 47.5 & 54.7 \\
Durall~\cite{Durall2020WatchYU} & 39.9 & 38.2 & 48.2 & 30.9 & 60.9 & 67.2 & 50.1 & 51.7 & 59.5 & 65.5 & 80.0 & 88.2 & \textbf{87.3} & 97.0 & 54.8 & 58.9 & 62.1 & 72.5 & 60.3 & 63.3 \\
Patchfor~\cite{Chai2020WhatMF}  & 68.0 & 92.9 & 97.1 & 100.0 & 97.8 & \textbf{99.9} & 93.6 & 98.2 & 97.9 & \textbf{100.0} & 99.6 & 100.0 & 66.8 & 68.1 & \textbf{97.6} & \textbf{99.8} & 92.7 & 99.8 & 90.1 & 95.4 \\
F3Net & 85.2 & 94.8 & 87.1 & 97.5 & 89.5 & 99.8 & 67.1 & 83.1 & 73.7 & 99.6 & 98.8 & 100.0 & 65.4 & 70.0 & 51.6 & 93.6 & 60.3 & 99.9 & 75.4 & 93.1 \\
SelfBland~\cite{Shiohara2022DetectingDW}  & 63.1 & 66.1 & 56.4 & 59.0 & 75.1 & 82.4 & 79.0 & 82.5 & 68.6 & 74.0 & 73.6 & 77.8 & 53.2 & 53.9 & 61.6 & 65.0 & 61.2 & 66.7 & 65.8 & 69.7 \\
GANDetection~\cite{Mandelli2022DetectingGI} & 57.4 & 75.1 & 67.9 & 100.0 & 67.8 & 99.7 & 67.6 & 92.4 & 67.7 & 99.3 & 60.9 & 86.2 & 69.6 & 83.5 & 66.7 & 90.6 & 69.6 & 97.2 & 66.1 & 91.6 \\
LGrad~\cite{Tan2023LearningOG}  & 68.6 & 93.8 & 69.9 & 89.2 & 50.3 & 54.0 & 71.1 & 82.0 & 57.5 & 67.3 & 89.1 & 99.1 & 78.5 & 86.0 & 78.0 & 87.4 & 54.8 & 68.0 & 68.6 & 80.8\\
Ojha~\cite{Ojha2023TowardsUF} & 78.5 & 98.3 & 72.0 & 98.9 & 77.6 & 99.8 & 77.6 & 98.9 & 77.6 & 99.7 & 78.2 & 98.7 & 85.2 & \textbf{98.1} & 77.6 & 98.7 & 74.2 & 97.8 & 77.6 & \textbf{98.8}\\
NPR~\cite{Tan2023RethinkingTU} &  83.0 & 96.2 & \textbf{99.0} & 99.8 & \textbf{98.7} & 99.0 & \textbf{94.5} & 98.3 & \textbf{98.6} & 99.0 & 99.6 & 100.0 & 79.0 & 80.0 & 88.8 & 97.4 & \textbf{98.0} & \textbf{100.0} & 93.2 & 96.6 \\
\rowcolor{lightgray} {\textbf{ADOF(ours)}} & \textbf{99.5} & \textbf{100.0} & 92.2 & \textbf{100.0} & 96.0 & 99.6 & 94.1 & \textbf{99.1} & 96.0 & 99.7 & \textbf{100.0} & \textbf{100.0} & 77.5 & 86.7 & 94.8 & 99.3 & 97.8 & 99.7 & \textbf{94.2} & 98.2 \\
\hline
\end{tabular}
\end{adjustbox}
\end{table}

\begin{table}[t!]
\centering
\caption{Evaluation results on the test set of DiffusionForensics dataset~\cite{Wang2023DIREFD}.}
\label{tab:table3}
\begin{adjustbox}{max width=\textwidth}
\begin{tabular}{l cc cc cc cc cc cc cc cc|cc}
\toprule
\multirow{2}{*}{\shortstack{\textbf{Method}}} & \multicolumn{2}{c}{\shortstack{\textbf{ADM}}} & \multicolumn{2}{c}{\shortstack{\textbf{DDPM}}} & \multicolumn{2}{c}{\shortstack{\textbf{IDDPM}}} & \multicolumn{2}{c}{\shortstack{\textbf{LDM}}} & \multicolumn{2}{c}{\shortstack{\textbf{PNDM}}} & \multicolumn{2}{c}{\shortstack{\textbf{VQ-Diffusion}}} & \multicolumn{2}{c}{\shortstack{\textbf{Stable}\\\textbf{Diffusion v1}}} & \multicolumn{2}{c}{\shortstack{\textbf{Stable}\\\textbf{Diffusion v2}}} & \multicolumn{2}{c}{\shortstack{\textbf{Mean}}} \\

\cline{2-19}
& \textbf{Acc.} & \textbf{A.P.} & \textbf{Acc.} & \textbf{A.P.} & \textbf{Acc.} & \textbf{A.P.} & \textbf{Acc.} & \textbf{A.P.} & \textbf{Acc.} & \textbf{A.P.} & \textbf{Acc.} & \textbf{A.P.} & \textbf{Acc.} & \textbf{A.P.} & \textbf{Acc.} & \textbf{A.P.} & \textbf{Acc.} & \textbf{A.P.} \\

\hline
CNNDetection~\cite{Wang2019CNNGeneratedIA}  & 53.9 & 71.8 & 62.7 & 76.6 & 50.2 & 82.7 & 50.4 & 78.7 & 50.8 & 90.3 & 50.0 & 71.0 & 38.0 & 76.7 & 52.0 & 90.3 & 51.0 & 79.8 \\
Frank~\cite{Frank2020LeveragingFA}  & 58.9 & 65.9 & 37.0 & 27.6 & 51.4 & 65.0 & 51.7 & 48.5 & 44.0 & 38.2 & 51.7 & 66.7 & 32.8 & 52.3 & 40.8 & 37.5 & 46.0 & 50.2 \\
Durall~\cite{Durall2020WatchYU}  & 39.8 & 42.1 & 52.9 & 49.8 & 55.3 & 56.7 & 43.1 & 39.9 & 44.5 & 47.3 & 38.6 & 38.3 & 39.5 & 56.3 & 62.1 & 55.8 & 47.0 & 48.3 \\
Patchfor~\cite{Chai2020WhatMF}   & 77.5 & 93.9 & 62.3 & 97.1 & 50.0 & 91.6 & 99.5 & 100.0 & 50.2 & 99.9 & 100.0 & 100.0 & 90.7 & 99.8 & 94.8 & 100.0 & 78.1 & 97.8\\
F3Net~\cite{Qian2020ThinkingIF} & 80.9 & 96.9 & 84.7 & 99.4 & 74.7 & 98.9 & 100.0 & 100.0 & 72.8 & 99.5 & 100.0 & 100.0 & 73.4 & 97.2 & 99.8 & 100.0 & 85.8 & 99.0 \\
SelfBland~\cite{Shiohara2022DetectingDW}   & 57.0 & 59.0 & 61.9 & 49.6 & 63.2 & 66.9 & 83.3 & 92.2 & 48.2 & 48.2 & 77.2 & 82.7 & 46.2 & 68.0 & 71.2 & 73.9 & 63.5 & 67.6 \\
GANDetection~\cite{Mandelli2022DetectingGI}  & 51.1 & 53.1 & 62.3 & 46.4 & 50.2 & 63.0 & 51.6 & 48.1 & 50.6 & 79.0 & 51.1 & 51.2 & 39.8 & 65.6 & 50.1 & 36.9 & 50.8 & 55.4 \\
LGrad~\cite{Tan2023LearningOG}   & 86.4 & 97.5 & \textbf{99.9} & 100.0 & 66.1 & 92.8 & 99.7 & 100.0 & 69.5 & 98.5 & 96.2 & 100.0 & 90.4 & 99.4 & 97.1 & 100.0 & 88.2 & 98.5 \\
Ojha~\cite{Ojha2023TowardsUF} & 78.4 & 92.1 & 72.9 & 78.8 & 75.0 & 92.8 & 82.2 & 97.1 & 75.3 & 92.5 & 83.5 & 97.7 & 56.4 & 90.4 & 71.5 & 92.4 & 74.4 & 91.7\\
NPR~\cite{Tan2023RethinkingTU}  & 88.6 & 98.9 & 99.8 & 100.0 & 91.8 & 99.8 & 100.0 & 100.0 & 91.2 & \textbf{100.0} & 100.0 & 100.0 & 97.4 & 99.8 & 93.8 & 100.0 & 95.3 & 99.8 \\
\rowcolor{lightgray} {\textbf{ADOF(ours)}} & \textbf{93.5} & \textbf{99.0} & 99.6 & \textbf{100.0} & \textbf{99.2} & \textbf{100.0} & 99.9 & \textbf{100.0} & \textbf{97.4} & 99.9 & 97.1 & 99.8 & \textbf{99.8} & \textbf{100.0} & \textbf{99.9} & \textbf{100.0} & \textbf{98.3} & \textbf{99.8} \\

\bottomrule
\end{tabular}
\end{adjustbox}
\end{table}

\begin{table}[t!]
\caption{Evaluation results on the diffusion test set of Ojha~\cite{Ojha2023TowardsUF}.}
\label{tab:table4}
\centering
\begin{adjustbox}{max width=\textwidth}
\begin{tabular}{l cc cc cc cc cc cc cc cc|cc}
\hline
\multirow{2}{*}{{\textbf{Method}}} & \multicolumn{2}{c}{\textbf{DALLE}} & \multicolumn{2}{c}{\textbf{Glide\_100\_10}} & \multicolumn{2}{c}{\textbf{Glide\_100\_27}} & \multicolumn{2}{c}{\textbf{Glide\_50\_27}} & \multicolumn{2}{c}{\textbf{ADM}} & \multicolumn{2}{c}{\textbf{LDM\_100}} & \multicolumn{2}{c}{\textbf{LDM\_200}} & \multicolumn{2}{c|}{\textbf{LDM\_200\_cfg}} & \multicolumn{2}{c}{\textbf{Mean}} \\
\cline{2-19}
& {\textbf{Acc.}} & {\textbf{A.P.}} & {\textbf{Acc.}} & {\textbf{A.P.}} & {\textbf{Acc.}} & {\textbf{A.P.}} & {\textbf{Acc.}} & {\textbf{A.P.}} & {\textbf{Acc.}} & {\textbf{A.P.}} & {\textbf{Acc.}} & {\textbf{A.P.}} & {\textbf{Acc.}} & {\textbf{A.P.}} & {\textbf{Acc.}} & {\textbf{A.P.}} & {\textbf{Acc.}} & {\textbf{A.P.}} \\

\hline
CNNDetection~\cite{Wang2019CNNGeneratedIA}  & 51.8 & 61.3 & 53.3 & 72.9 & 53.0 & 71.3 & 54.2 & 76.0 & 54.9 & 66.6 & 51.9 & 63.7 & 52.0 & 64.5 & 51.6 & 63.1 & 52.8 & 67.4 \\
Frank~\cite{Frank2020LeveragingFA}   & 57.0 & 62.5 & 53.6 & 44.3 & 50.4 & 40.8 & 52.0 & 42.3 & 53.4 & 52.5 & 56.6 & 51.3 & 56.4 & 50.9 & 56.5 & 52.1 & 54.5 & 49.6 \\
Durall~\cite{Durall2020WatchYU}  & 55.9 & 58.0 & 54.9 & 52.3 & 48.9 & 46.9 & 51.7 & 49.9 & 40.6 & 42.3 & 62.0 & 62.6 & 61.7 & 61.7 & 58.4 & 58.5 & 54.3 & 54.0 \\
Patchfor~\cite{Chai2020WhatMF}   & 79.8 & {99.1} & 87.3 & 99.7 & 82.8 & 99.1 & 84.9 & 98.8 & 74.2 & 81.4 & 95.8 & 99.8 & 95.6 & 99.9 & 94.0 & 99.8 & 86.8 & 97.2 \\
F3Net~\cite{Qian2020ThinkingIF}  & 71.6 & 79.9 & 88.3 & 95.4 & 87.0 & 94.5 & 88.5 & 95.4 & 69.2 & 70.8 & 74.1 & 84.0 & 73.4 & 83.3 & 80.7 & 89.1 & 79.1 & 86.5 \\
SelfBland~\cite{Shiohara2022DetectingDW}    & 52.4 & 51.6 & 58.8 & 63.2 & 59.4 & 64.1 & 64.2 & 68.3 & 58.3 & 63.4 & 53.0 & 54.0 & 52.6 & 51.9 & 51.9 & 52.6 & 56.3 & 58.7 \\
GANDetection~\cite{Mandelli2022DetectingGI}  & 67.2 & 83.0 & 51.2 & 52.6 & 51.1 & 51.9 & 51.7 & 53.5 & 49.6 & 49.0 & 54.7 & 65.8 & 54.9 & 65.9 & 53.8 & 58.9 & 54.3 & 60.1 \\
LGrad~\cite{Tan2023LearningOG}    & 88.5 & 97.3 & 89.4 & 94.9 & 87.4 & 93.2 & 90.7 & 95.1 & \textbf{86.6} & \textbf{100.0} & 94.8 & 99.2 & 94.2 & 99.1 & 95.9 & 99.2 & 90.9 & 97.2 \\
Ojha~\cite{Ojha2023TowardsUF}   & 89.5 & 96.8 & 90.1 & 97.0 & 90.7 & 97.2 & 91.1 & 97.4 & 75.7 & 85.1 & 90.5 & 97.0 & 90.2 & 97.1 & 77.3 & 88.6 & 86.9 & 94.5 \\
NPR~\cite{Tan2023RethinkingTU}   & {94.5} & \textbf{99.5} & 98.2 & 99.8 & 97.8 & 99.7 & 98.2 & 99.8 & 75.8 & 81.0 & \textbf{99.3} & 99.9 & \textbf{99.1} & 99.9 & \textbf{99.0} & 99.8 & \textbf{95.2} & 97.4 \\
RINE~\cite{koutlis2024leveraging}   & \textbf{95.0} & 99.5 & 90.7 & 99.2 & 88.9 & 99.1 & 92.6 & 99.5 & 76.1 & 96.6 & {98.7} & 99.9 & {98.3} & 99.9 & 88.2 & 98.7 & {91.1} & 99.0 \\
\rowcolor{lightgray} {\textbf{ADOF(ours)}} & 92.1 & 98.3 & \textbf{98.6} & \textbf{100.0} & \textbf{98.7} & \textbf{100.0} & \textbf{98.4} & \textbf{99.9} & 75.9 & 87.6 & 98.8 & \textbf{100.0} & 98.6 & \textbf{99.9} & 98.5 & \textbf{99.9} & 94.9 & \textbf{98.2} \\

\hline
\end{tabular}
\end{adjustbox}
\end{table}


\subsection{Computational Efficiency Evaluation}
We conduct a comparative analysis of several representative methods based on their computational efficiency and resource requirements. Specifically, we evaluate and compare the following metrics:
\begin{figure}
\includegraphics[width=\textwidth]{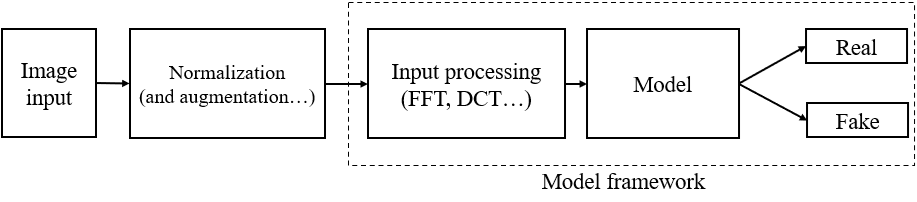}
\caption{Standard pipeline for Image Synthesis Methods}
\label{figure_pine_line_1}
\end{figure}
\begin{itemize}
\item \textbf{Number of Parameters:} We quantify the total number of parameters for each model to assess its complexity.
\item \textbf{Input Processing Time:} We measure the time required for processing images before they are fed into the model, where these processing steps are tailored to the specific method used (See Fig.~\ref{figure_pine_line_1}).
\item \textbf{Inference Time:} We record the time taken for the model to process an image and produce a result.
\item \textbf{FLOPs (Floating Point Operations Per Second):} We leveraged the \texttt{fvcore} library to estimate the FLOPs required by each model during inference, providing valuable insights into their computational demands.
\end{itemize}
\begin{table}[t!]
	\centering
	\caption{Resource usage and performance of synthetic image detection methods on the DiffusionForensics~\cite{Wang2023DIREFD}. The method marked with \textsuperscript{$\dagger$} indicates it was trained on this dataset.}
	\label{table_generative_models}
    \begin{adjustbox}{max width=\textwidth}
	\setlength{\tabcolsep}{12pt} 
	\renewcommand{\arraystretch}{1.1} 
	\begin{tabular}{l c c c c c}
		\toprule
		\multirow{2}{*}{\textbf{Method}} &
		\multirow{2}{*}{\textbf{Parameters}} &
		\multirow{2}{*}{\begin{tabular}[c]{@{}c@{}}\textbf{Processing} \\ \textbf{(ms)}\end{tabular}} &
		\multirow{2}{*}{\begin{tabular}[c]{@{}c@{}}\textbf{Inference Time}\\ \textbf{(ms)}\end{tabular}} &
		\multirow{2}{*}{\textbf{FLOPs}} &
		\multirow{2}{*}{\begin{tabular}[c]{@{}c@{}}\textbf{Means} \\ \textbf{(acc/ap)} \end{tabular}} \\
		&   &   &   &   &    \\ \hline
		LGrad~\cite{Tan2023LearningOG} & {$25.56 \times 10^6$} & {11.6} & {4.81} & {$4.12 \times 10^9$} & {88.2/98.5}\\ 
		DIRE\textsuperscript{$\dagger$}~\cite{Wang2023DIREFD} & {$25.56 \times 10^6$} & {4,502.7} & {4.81} & {$4.12 \times 10^9$} & {97.9/\textbf{100}}\\ 
		Ojha~\cite{Ojha2023TowardsUF} & {$427.62 \times 10^6$} & {None} & {29.19} & {$77.83 \times 10^9$} & {74.4/91.7}\\ 
		\rowcolor{lightgray} \textbf{ADOF(ours)} & {\textbf{$1.44 \times 10^6$}} & 0.40 & \textbf{2.43} & {$\mathbf{1.74 \times 10^9}$} & {\textbf{98.3}/99.8}\\ \bottomrule
	\end{tabular}
    \end{adjustbox}
\end{table}


\textcolor{black}{Our method requires substantially fewer parameters and FLOPs while achieving faster inference and the highest mean accuracy (98.3\%) compared to existing methods, including the DIRE~\cite{Wang2023DIREFD} (see Table~\ref{table_generative_models}), which is trained on the same dataset but does not achieve comparable performance. This demonstrates its superior performance in synthetic image detection.}

\section{Conclusion}
In this paper, we proposed a simple yet highly effective filter, namely ADOF, for capturing pixel-level variations. By treating an image as a discrete digital signal, this method eliminates the average components of the signal. These components typically carry semantic information, which is less helpful for distinguishing between real and synthetic images compared to the subtle traces that the proposed filter is designed to detect. Experimental results indicates that our proposed method significantly reduces model complexity while enhancing both accuracy and generalization, even on previously unseen data.

%
\section*{Acknowledgment}
This research is funded by Vietnam National University - Ho Chi Minh City (VNU-HCM) under Grant Number C2024-18-25. 

\bibliographystyle{splncs04}
\bibliography{sections/bibliography} 
\end{document}